# Semantic2Graph: Graph-based Multi-modal Feature Fusion for Action Segmentation in Videos

Junbin Zhang[1], Pei-Hsuan Tsai[2*] and Meng-Hsun Tsai[1, 3]

[1]Department of Computer Science and Information Engineering, National Cheng Kung University, Tainan, 701, Taiwan.
[2*]Institute of Manufacturing Information and Systems, National Cheng Kung University, Tainan, 701, Taiwan
[3]Department of Computer Science, National Yang Ming Chiao Tung University, Hsinchu, 300, Taiwan

*Corresponding author(s). E-mail(s): phtsai@mail.ncku.edu.tw;
Contributing authors: p78083025@ncku.edu.tw; tsaimh@cs.nycu.edu.tw;

**Abstract**

Video action segmentation have been widely applied in many fields. Most previous studies employed video-based vision models for this purpose. However, they often rely on a large receptive field, LSTM or Transformer methods to capture long-term dependencies within videos, leading to significant computational resource requirements. To address this challenge, graph-based model was proposed. However, previous graph-based models are less accurate. Hence, this study introduces a graph-structured approach named *Semantic2Graph*, to model long-term dependencies in videos, thereby reducing computational costs and raise the accuracy. We construct a graph structure of video at the frame-level. Temporal edges are utilized to model the temporal relations and action order within videos. Additionally, we have designed positive and negative semantic edges, accompanied by corresponding edge weights, to capture both long-term and short-term semantic relationships in video actions. Node attributes encompass a rich set of multi-modal features extracted from video content, graph structures, and label text, encompassing visual, structural, and semantic cues. To synthesize this multi-modal information effectively, we employ a graph neural network (GNN) model to fuse multi-modal features for node action label classification. Experimental results demonstrate that Semantic2Graph outperforms state-of-the-art methods in terms of performance, particularly on benchmark datasets such as GTEA and 50Salads. Multiple ablation experiments further validate the effectiveness of semantic features in enhancing model performance. Notably, the inclusion of semantic edges in Semantic2Graph allows for the cost-effective capture of long-term dependencies, affirming its utility in addressing the challenges posed by computational resource constraints in video-based vision models.

## 1 Introduction

Video action segmentation stands as a pivotal technology extensively employed across diverse applications and has attracted a great deal of study interest in recent years. Owing to the fruitful progress of deep learning in computer vision tasks, early methods utilize video-based vision models for video action segmentation [1]. In video-based vision models, video is viewed as a sequence of RGB frames [2]. The video-based vision models frequently rely on expanding sliding windows (receptive fields) [1, 3], increasing network depth, introducing attention mechanisms, stacking modules, LSTM or Transformer methods to extract spatio-temporal varying local and global features from videos to represent complex relations in videos [3, 4]. However, these dependency modeling approaches, while effective, comes at a substantial computational cost [1].

With the advances in Graph Neural Networks (GNNs), numerous graph models have been implemented for video action segmentation and recognition over the last few years [1, 3, 5, 6]. Most video-based graph models are *skeleton-based*, which extract skeleton graphs from video frames relying on human pose information instead of RGB pixel data [2, 7-9]. A video is represented as a sequence of skeleton graphs. However, skeleton-based method degrades visual features and is only appropriate for specific scenes with skeleton information. Consequently, several studies [1, 3, 6, 10] transform a video (or video clip) into a graph in which each node represents a video frame (or video clip). These methods are *graph-based* in this paper. For example, Hamza Khan et al. [5] uses graph convolutional network to learn the visual features of frames and the relationship between neighbor frames for timestamp-supervised action segmentation. Dong Wang et al. [6] proposed Dilated Temporal Graph Reasoning Module (DTGRM) to construct a multi-layer detailed temporal graph of the video to capture and model the temporal relationships and dependencies in the video. They construct S-Graph (Similarity Graph) and L-Graph (Learned Graph) for each frame. The attribute of each node is frame-wise feature that extracted by I3D. Graph-based methods have larger receptive fields and lower costs to capture long-term dependencies [1, 3, 6], and are also able to capture non-sequential temporal relations. Thus, most graph-based models can address the computationally expensive problem of visual models capturing long-term dependencies in videos. In recent years, many works [1, 2, 9, 11-13] have demonstrated that multi-modal features substantially improve the accuracy of results [9, 14]. However, compared to vision models, existing graph-based methods are less accurate. Because they only utilize the visual features of frames as node attributes and ignore other modal features (such as label text), which results in suboptimal results. Therefore, multi-modal features are adopted to facilitate video reasoning, especially for natural language supervised information [1, 4, 15, 16] in this paper. Hence, the motivation of this study is to design a graph-based model with higher accuracy.

Our approach is to leverage the powerful relational modeling of graphs to reduce computational and multi-modal feature fusion further improves graph model performance. The graph-structured representation of video provides valuable information about the temporal relation between different frames or video segments. The graph structure can capture both temporal coherence and action order, which allows us to model the sequential nature of actions in videos. Hence, in this paper, we introduce a graph-based method named *Semantic2Graph*, to turn the video action segmentation into node classification. The explicit goal of Semantic2Graph is to reduce the computational demands associated with video-based vision models. To this end, we construct a graph structure of video at the frame-level. Adding temporal edges, semantic edges, and self-loop edges to capture both long-term and short-term relations in the video frame sequences. Moreover, node attributes encompass a rich set of multi-modal features, include visual, structural, and semantic cues. Structural features provide important action order information. The incorporation of structural features enables our approach to model the dynamic evolution of actions over time and helps to accurately segment actions and recognize boundaries. Semantic features provide a higher-level understanding of actions occurring in videos. Furthermore, semantic features can assist in handling challenging scenarios, such as ambiguous or rare actions, by providing additional cues for disambiguation. Node neighborhood information is reflected through structural features. Notably, semantic features are the text embedding of sentences encoded by a visual language pre-trained model [4, 15, 16], and the sentence is label-text extended using prompt-based method [4, 15, 16]. A GNNs model, that is a spatial-based graph convolutional neural network (GCN) model suitable for processing directed graphs [17, 18], is employed for multi-modal feature fusion to predict node action labels. Experimental results demonstrate that our approach outperforms state-of-the-art methods in terms of performance. Our contributions are summarized as follows:

- We present a small-scale and flexible graph model for video action segmentation.
- We describe a method for constructing the graph of videos and design three kinds of edges to model fine-grained relations in videos. Especially, the semantic edges reduce the cost of capturing long-term temporal relations.
- We introduce multi-modal features combining visual, structural, and semantic features as node attributes. In particular, the semantic features of textual prompt-based are more effective than only label words for enhance semantic content.
- We use a GNNs model to fuse visual, structural, and semantic features, and demonstrate that middle-level and high-level features are the key to further model performance improvement.
- Experimental results demonstrate that our approach holds promise in not only preserving the accuracy of action segmentation but also in making the process more resource-efficient.

In the following sections, we will elaborate on the details of our graph-based approach and present empirical evidence of its efficacy. In the next section, we first introduce how we construct the graph model.

## 2 Our approach

This work aims to address the challenges of video action segmentation and recognition utilizing graph models. To do this, we present a graph-based method named *Semantic2Graph*.

### 2.1 Notation

Let V = {$v_1$, ..., $v_N$} denote the set of untrimmed videos. The $i$-th video $v_i$ = {$f_t \in \mathbb{R}^{H \times W \times C}$}$_{t=1}^{T_i}$ ∈V contains $T_i$ frames, where $f_t$ denote the $t$-th frame with height $H$, width $W$ and channel $C$ of a video [1]. Let a directed graph $\mathcal{DiG}(\mathcal{V}, \mathcal{E})$ with $T$ nodes denote a graph-structured of video. In $\mathcal{DiG}(\mathcal{V}, \mathcal{E})$, $\mathcal{V} \in \mathbb{R}^{1 \times T}$ is a set of nodes and a node $v_t \in \mathcal{V}$. $\mathcal{E}$ is a set of edges and an edge $e_t = (v_t, v_{t'}) \in \mathcal{E}$, where $(v_t, v_{t'})$ means the node $v_t$ goes into $v_{t'}$. $w$ is weight of an edge, $w \in \{\gamma, 1\}$, $0 \leq \gamma < 1$. The node features of the graph $\mathcal{DiG}$ are X = {$X_1$, ..., $X_T$} ∈ $\mathbb{R}^{T \times D}$, here D is the dimension of the node feature. $\mathcal{Y} = \{y_t\}_{t=1}^T$ is a set of node labels. G = {$\mathcal{DiG}_1, ..., \mathcal{DiG}_N$} denote the set of graphs.

### 2.2 General scheme of our approach

Fig. 1 illustrates the Semantic2Graph pipeline, which is divided into the following (a) to (e) steps.

Step (a) is visual feature extraction, as shown in Fig. 1(a). The video is split into frames. Semantic2Graph utilizes a 3D convolutional network to extract visual features from video frames for each frame. Visual features are the most fundamental and important features.

Step (b) is graph construction, as shown in Fig. 1(b). Its inputs are frame-set and visual features from step (a) and the detailed video labels. Each frame represents a node. The attribute and label of a node are the visual feature and the action label of the frame, respectively. Nodes with different labels are distinguished by their color. There are three types of edges to maximize the preservation of relations in video frames. The detailed edge construction is described in Section 2.3. The output is a directed graph.

Step (c) is structure embedding which is to encode the neighborhood information of nodes as structural features, as shown in Fig. 1(c). The neighborhood information is provided by the directed graph from step (b).

Step (d) is semantic embedding, as shown in Fig. 1(d). The label text of video frames is expanded into sentences by prompt-based CLIP. Sentences are encoded by a text encoder to obtain semantic features.

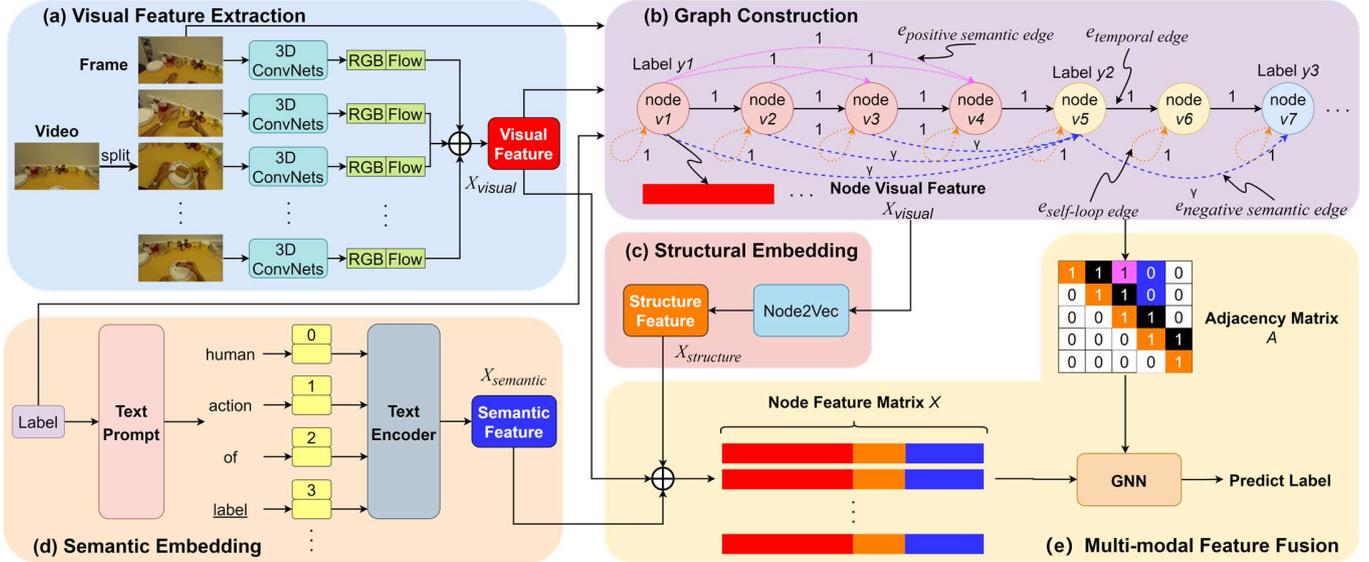

**Fig. 1** Overview of the pipeline for Semantic2Graph. (a) Extracting visual features from video frames. (b) An instance of a video-based graph. (c) Encoding node neighborhood information as structural features. (d) Encoding the label text to obtain semantic features. (e) A GNNs is trained to fusion multi-modal features to predict node labels.

Step (e) is multi-modal feature fusion, as shown in Fig. 1(e). Its inputs include visual features from step (a), structural features from step (c), semantic features from step (d), and a directed graph from step (b). A weighted adjacency matrix is generated from the directed graph. The multi-modal feature matrix is obtained by concatenating visual features, structural features, and semantic features. A GNNs is selected as the backbone model to learn multi-modal feature fusion. The output is the node's predicted action label.

## 2.3 Graph construction of video

This section is to describe how a video is transformed into our defined graph. The defined graph is a directed graph $\mathcal{DiG}(\mathcal{V}, \mathcal{E})$ composed of a set of nodes $\mathcal{V}$ and a set of edges $\mathcal{E}$. Compared with undirected graph, directed graph are more conducive to modeling temporal or order relations [18].

### 2.3.1 Node

The input is a set of untrimmed videos. A frame $f_t$ is the frame at time slot $t$ in the video. Video $v$ is a set of $f_t$, where the set contains $T$ frames. As shown in Fig. 1 (b) that video $v$ is represented by a directed graph $\mathcal{DiG}(\mathcal{V}, \mathcal{E})$. A frame $f$ in a video is represented by a node $v$ in the directed graph. In other words, a directed graph with $T$ nodes represents a video with $T$ frames. Directed graphs can explicitly model temporal relations in videos, and can also model action-order relations. Especially irreversible action pairs, such as door opening-closing, egg breaking-stirring.

For each node, there is a label $y$ converted from the label of the represented frame. Without loss of generality, we assume that the video has pre-obtained frame-level action annotations by some methods, such as frame-based methods [2, 6, 14], segment-based methods, or proposal-based methods [1, 3, 8, 19]. In addition, each node has three attributes that are multi-modal features extracted from frames. The details of extracting frame features are subsequently specified in Section 2.4.

### 2.3.2 Edge

Videos contain not only rich sequential dependencies between frames, but also potential semantic relations, such as captions, context, action interactions, object categories, and more. These dependencies are crucial for a comprehensive understanding of video [19].

To enhance the graph representation learning of videos, there are two types of edges in the directed graph, temporal edges, and semantic edges, as shown in Fig. 1 (b). In addition, to enhance the complexity of a graph and preserve the features of the node itself, another type of edge self-loop is added to the directed graph [19]. Each edge has a weight value determined by the type of edge.

*Temporal edge.* It is the baseline edge of the directed graph to represent the sequential order of frames. Intuitively, it reflects the temporal adjacency relation between two neighbor nodes [1, 6, 19, 20]. In general, the weight of the temporal edge is set to 1. The temporal edge is formulated as

$$e_{temporal} = (v_t, v_{t+1}), \quad t \in [1, T-1] \qquad (1)$$
$$w_{temporal} = 1 \qquad (2)$$

*Semantic edge.* It is to model the potential semantic relations of videos [1]. Semantic edges are categorized into two types, namely, *positive semantic edge* and *negative semantic edge*. Positive semantic edge is to group nodes with identical labels while negative semantic edge is to distinguish nodes with different labels. A simple rule to construct semantic edges is to add either positive or negative edges between any two nodes based on their labels. However, the size of edges is $O(\mathcal{V}^2)$ resulting in the consumption of memory space and computing time. We here propose a semantic edge construction method to optimize the directed graph with minimum edges and sufficient semantic relations of video.

Each node has a sequence identifier which is the same as the ID of the representative frame. The smaller the ID number, the earlier the frame in a video. For each node, $v_i$ adds a positive edge to the following nodes $v_j$ whose $j > i$ until the label of $v_j$ is different from the label of $v_i$ and add a negative edge to $v_j$. Positive semantic edge and negative semantic edge are defined as

$$e_{positive\_semantic} = (v_i, v_j), \quad \begin{cases} i,j \in [1,T] \text{ and } j > i \\ y_i = y_j \end{cases}$$
$$w_{positive\_semantic} = 1, \tag{3}$$

$$e_{negative\_semantic} = (v_i, v_j), \quad \begin{cases} i,j \in [1,T] \text{ and } j > i \\ y_i \neq y_j \end{cases}$$
$$w_{negative\_semantic} = \gamma, \tag{4}$$

*Self-loop edge.* It is referring to the settings of the MIGCN model that each node adds a self-loop edge with a weight of 1. During message aggregation, the self-loop edge maintains the node's information [19]. The self-loop edge is formulated as

$$e_{self-loop} = (v_t, v_t), \quad t \in [1,T] \tag{5}$$
$$w_{self-loop} = 1 \tag{6}$$

## 2.4 Multi-modal features

Video contents are extracted as features and added to nodes in the graph. They are visual, structure, and semantic features, belonging to low-level, middle-level, and high-level features [14], respectively, as shown in Fig. 1.

*Visual feature.* Visual features are the most fundamental and important features. It captures the appearance information, such as color, texture, and motion, which are essential for understanding actions in videos. In our method, it is an image embedding containing RGB and optical-flow information of each frame in a video, as shown in Fig. 1 (a). Optical flow estimation techniques to extract motion features aim at capturing temporal dynamics. Many visual feature extractors have been developed, such as I3D, C3D, ViT, etc. [1, 20]. In this study, we use I3D to extract visual features. The I3D feature extractor takes a video $v_i$ input and outputs two tensors with 1024-dimensional features: for RGB and optical-flow streams. Visual features concatenate RGB and optical-flow feature tensors. Visual feature is represented by $X_{visual} \in \mathbb{R}^{T \times D_{visual}}$, $D_{visual} = 2048$.

$$X_{visual,i} = \text{I3D}(v_i) \tag{7}$$

*Structure Feature.* It is the node embedding transferring features into a low-dimensional space with minimum loss of information of the neighborhood nodes. General structure features are the structure of a graph. In this paper, structure features reflect the structural properties of a graph.

The graph-structured representation of video provides valuable information about the temporal relation between different frames or video segments. The graph structure can capture both temporal coherence and action order, which allows us to model the sequential nature of actions in videos. The structural properties are relational properties of a video, including intrinsic sequence-structure properties and semantic properties. They are preserved by our designed temporal edge and semantic edge. See Section 2.3.2 for details. Structural features provide important action order information. The combination of structural features enables our method to model the dynamic evolution of actions over time and helps to accurately segment actions and recognize boundaries.

In computer vision, most of the existing methods [15, 19, 21] use Recurrent Neural Networks (RNN) sequence models such as GRU, BiGRU [19], LSTM or Transformer [15, 21] to capture the intrinsic sequence-structure properties of videos, and then to reason the mapping relation between frame and action. However, the methods mentioned above are not suitable for graphs. The reason is that they are not good at dealing with non-Euclidean forms of graphs.

For a graph, the structure features are obtained by using a node embedding algorithm. There are many node embedding algorithms, such as DeepWalk for learning the similarity of neighbors, LINE for learning the similarity of first-order and second-order neighbors, and node2vec [22] for learning the similarity of neighbors and structural similarity.

In this paper, node2vec is used to get the structural features, as shown in Fig. 1 (c). The structure feature is represented by $X_{structure} \in \mathbb{R}^{T \times D_{structure}}$, where dimension $D_{structure} = 128$.

$$X_{structure} = \text{node2vec}(\mathcal{D}i\mathcal{G}(\mathcal{V}, \mathcal{E})) \tag{8}$$

where the node attributes of the graph $\mathcal{D}i\mathcal{G}(\mathcal{V}, \mathcal{E})$ are only visual features.

*Semantic feature.* It is the language embedding of each frame of a video, such as textual prompt [16] or the semantic information of label-text [4], as shown in Fig. 1 (d). Semantic features provide a higher-level understanding of actions occurring in videos. Furthermore, semantic features can assist in handling challenging scenarios, such as ambiguous or rare actions, by providing additional cues for disambiguation. CLIP [15] and ActionCLIP [4] are common approaches to getting semantic features. In this study, we use ActionCLIP.

Following the ActionCLIP model proposed by Wang et al. [4], based on the filling locations, the filling function $\mathcal{T}$ has the following three varieties:
- Prefix prompt [4]: *label, a video of action;*
- Cloze prompt [4]: *this is label, a video of action;*
- Suffix prompt [4]: *human action of label.*

ActionCLIP uses the label-text to fill the sentence template $Z = \{z_1, ..., z_k\}$ with the filling function $\mathcal{T}$ to obtain the prompted textual [4].

$$\mathcal{Y}' = \mathcal{T}(\mathcal{Y}, Z) \tag{9}$$

Compared with only label words, the prompt-based method expands the label-text, which enhances the semantic features. A text encoder encodes is to obtain semantic features, which are used as language supervision information to improve the performance of vision tasks. The semantic feature is represented by $X_{semantic} \in \mathbb{R}^{T \times D_{semantic}}$, where dimension $D_{semantic} = 512$.

$$X_{semantic} = \text{Text\_Encoder}(\mathcal{Y}') \tag{10}$$

As shown in Fig. 1 (e) that the multi-modal feature of the node is

$$X = X_{visual} \parallel X_{structure} \parallel X_{semantic} \tag{11}$$

where || represents a concatenation operation.

## 2.5 Graph construction algorithms

Here is the algorithm pseudo code to describe how to create a directed graph from a video. In Algorithm 1, the input is a video $v$ and its set of action labels $\mathcal{Y}$, and the output is a directed graph. It is divided into 7 steps as follows:

*Step 1.* Initialization (see lines 1-2). Create an empty directed graph $\mathcal{DiG}(\mathcal{V}, \mathcal{E})$ consisting of node set $\mathcal{V}$ and edge set $\mathcal{E}$. Split the video $v$ into a set of frames $\{f_1, ..., f_T\}$.

*Step 2.* Create nodes (see lines 3-5). Creates a node $v_i$ based on frame-level. Node $v_i$ represents video frame $f_i$, so the label of node $v_i$ is the label $y_i$ of video frame $f_i$. Then add node $v_i$ to a node-set $\mathcal{V}$.

*Step 3.* Create temporal edges (see lines 6-9). Traverse the node set $\{v_2, ..., v_T\}$. Then a temporal edge $(v_{i-1}, v_i)$ with weight 1 from node $v_{i-1}$ to node $v_i$ is added to the edge set $\mathcal{E}$.

---

**Algorithm 1** Video to a Directed Graph

**Input**: Load a video $v \in V$ and a ground_truth_labels $\mathcal{Y} = \{y_1, ..., y_T\}$

**Output**: $\mathcal{DiG}(\mathcal{V}, \mathcal{E})$

    // Step 1: Initialization
1:   Create an empty directed graph $\mathcal{DiG}(\mathcal{V}, \mathcal{E})$
2:   $v$ split set of frames $\{f_1, ..., f_T\}$
    // Step 2: Create nodes
3:   **for** $i \in [1, T]$ **do**
4:     Create note $v_i$ and $\mathcal{V} \leftarrow$ add node $v_i$
5:   **end for**
    // Step 3: Create temporal edges
6:   **for** $i \in [2, T]$ **do**
7:     $\mathcal{E} \leftarrow$ add temporal edge $e_{temporal} = (v_{i-1}, v_i)$ and $w_{temporal} = 1$
8:     **end if**
9:   **end for**
    // Step 4: Create positive semantic edges
10:  node_group = $\{v_1\}$
11:  **for** $i \in [2, T]$ **do**
12:    **if** $y_i == y_{i-1}$ **then**
13:      **for** $v_j$ in node_group **do**
14:        **if** $j < i-1$ **then**
15:          $\mathcal{E} \leftarrow$ add positive semantic edge $e_{positive\_semantic} = (v_j, v_i)$ and $w_{positive\_semantic} = 1$
16:        **end if**
17:      **end for**
18:    **else** node_group = {}
19:    **end if**
20:    node_group $\leftarrow$ add $v_i$
21:  **end for**
    // Step 5: Create negative semantic edges
22:  node_group = $\{v_1\}$
23:  **for** $i \in [2, T]$ **do**
24:    **if** $y_i \neq y_{i-1}$ **then**
25:      **for** $v_j$ in node_group **do**
26:        **if** $i-j \neq 1$ **then**
27:          $\mathcal{E} \leftarrow$ add negative semantic edge $e_{negative\_semantic} = (v_j, v_i)$ and $w_{negative\_semantic} = \gamma$
28:        **end if**
29:      **end for**
30:      node_group = {}
31:    **end if**
32:    node_group $\leftarrow$ add $v_i$
33:  **end for**
    // Step 6: Create self-loop edges
34:  **for** $i \in [1, T]$ **do**
35:    $\mathcal{E} \leftarrow$ add self-loop edge $e_{self-loop} = (v_i, v_i)$ and $w_{self-loop} = 1$
36:  **end for**
    // Step 7: Get node attributes
37:  $X_{visual} = \{x_1, ..., x_T\} \leftarrow$ GetVisualFeatures($\{f_1, ..., f_T\}$)
38:  $\mathcal{V}$.attribute $\leftarrow X_{visual}$
39:  $X_{structure} \leftarrow$ GetStructureFeatures($\mathcal{DiG}(\mathcal{V}, \mathcal{E})$)
40:  $X_{semantic} \leftarrow$ GetSemanticFeatures($\mathcal{Y}$)
41:  $\mathcal{V}$.attribute $\leftarrow X = X_{visual} || X_{structure} || X_{semantic}$
42:  **save** $\mathcal{DiG}(\mathcal{V}, \mathcal{E})$

---

*Step 4.* Create positive semantic edges (see lines 10-21). First, create a node group *node_group* whose initial value is node $v_1$. Then, traverse the node set $\{v_2, ..., v_T\}$. If the label $y_i$ of the current node $v_i$ is the same as the label $y_{i-1}$ of the previous node $v_{i-1}$, then further traverse the node group. If node $v_j$ in *node_group* is not a neighbor of node $v_i$ (that is, $j \neq i-1$), then edge set $\mathcal{E}$ is added with a positive semantic edge $(v_j, v_i)$ with weight 1 from node $v_j$ to node $v_i$. Otherwise, *node_group* is emptied. Finally, node $v_i$ is added to *node_group*.

*Step 5.* Create negative semantic edges (see lines 21-33). Negative semantic edges and positive semantic edges have similar execution procedures. The difference is as follows: In line 26, only when the label $y_i$ of the current node $v_i$ is different from the label $y_{i-1}$ of the previous node $v_{i-1}$, the node group is further traversed. Except adjacent nodes, all nodes in *node_group* construct a negative semantic edge with node $v_i$ respectively. See line 28, edge set $\mathcal{E}$ is added with a negative semantic edge $(v_j, v_i)$ with weight $\gamma$ from node $v_j$ to node $v_i$. Then *node_group* is emptied.

*Step 6.* Create self-loop edges (see lines 34-36). Traverse the node set $\{v_1, ..., v_T\}$, and add a self-loop edge $(v_i, v_i)$ with weight 1 to the node set $\mathcal{E}$ for each node $v_i$.

*Step 7.* Get node attributes (see lines 37-41). First, extract visual features $X_{visual}$ (RGB and optical flow) from a set of frames $\{f_1, ..., f_T\}$ using a visual feature extractor. Then, the visual features $X_{visual}$ are used as node attributes of the directed graph $\mathcal{DiG}(\mathcal{V}, \mathcal{E})$. On $\mathcal{DiG}(\mathcal{V}, \mathcal{E})$, a node embedding model is used to encode the structural features $X_{structure}$ of node neighborhoods. Semantic features $X_{semantic}$ are obtained by using a textual prompt model to encode the label text $\mathcal{Y}$ of nodes. Concatenating $X_{visual}$, $X_{structure}$ and $X_{semantic}$ compositional multi-modal features $X$ as new attributes of nodes.

Finally, save the directed graph $\mathcal{DiG}(\mathcal{V}, \mathcal{E})$ with nodes, node attributes, node labels, edges, and edge weights transferred from video $v$.

An instance of the design of edges according to Algorithm 1 as illustrated in Fig. 2. It is worth noting that for adjacent nodes, according to the definitions of temporal and semantic edges in Section 2.3. A temporal edge and a positive semantic edge or a negative semantic edge should be added between them. But in

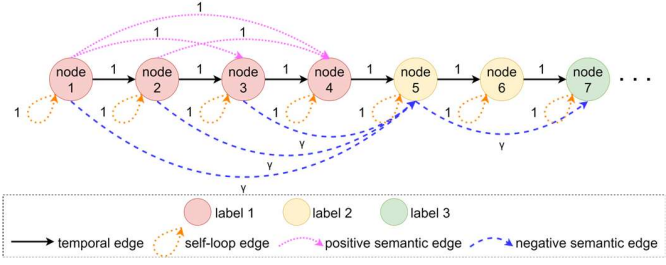

**Fig. 2** An instance of an edge in a directed graph. The color of the node represents different labels. Positive semantic edges (pink dashed lines) and negative semantic edges (blue dashed lines) are examples of semantic edges. The edge's value indicates its weight.

Algorithm 1 (see lines 6 and 7), only the temporal edges that are used to save the time series information of the video are added. See line 14, the purpose of setting the condition $j < i-1$ is to avoid adding positive semantic edges ($v_{i-1}, v_i$). See line 26, the purpose of setting the condition $i-j \neq 1$ is to avoid adding negative semantic edges ($v_{i-1}, v_i$). As shown in Fig. 2, nodes 1 and 2, nodes 2 and 3, nodes 3 and 4, nodes 4 and 5, and nodes 5 and 6 have only temporal edges.

For non-adjacent nodes, a positive semantic edge and a negative semantic edge should be added between them. But in Algorithm 1 (see lines 12 to 19), only non-adjacent nodes with the same label are added positive semantic edges. See lines 23 to 27, only non-adjacent nodes with the different label are added negative semantic edges. As shown in Fig. 2, nodes 1 and 3, nodes 1 and 4 and nodes 2 and 4 is positive semantic edges. While nodes 1 and 5, nodes 2 and 5, nodes 3 and 5 and nodes 5 and 7 is negative semantic edges.

For human-annotated videos, two consecutive frames may have different labels despite their similar visual features. To enhance the boundaries of semantic labels in the graph, the weight of positive semantic edges is set to 1 and the weight of negative semantic edges is set to $\gamma$. In Fig. 2, positive semantic edges enhance the semantic relation that node 4 belongs to the same label as nodes 1, 2, and 3. Negative semantic edges enhance the semantic relation between node 5 and nodes 1, 2, and 3 belonging to different labels. In summary, semantic edges also help in label class prediction for two adjacent nodes whose true labels are not identical.

In a graph with only temporal edges, a node has only two neighbors, so the message aggregation within 1 hop is limited. Moreover, if the nodes want to aggregate messages from more nodes or more distant nodes, the model needs to go through multi-hop and expensive computation. Conversely, semantic edges allow the model to be implemented within 1 hop. Hence, semantic edges significantly reduce the computational cost of message aggregation in GNNs [1].

## 2.6 Graph-based fusion model

In this paper, we treat video action segmentation and recognition as a node classification problem on graphs. We need to define a weighted adjacency matrix $\mathcal{A}$ with weight that relies on the three types of edges and their weights proposed in Section 2.3 [23]. Calculated as follows

$$\mathcal{A}_{i,j} = \begin{cases} w_{temporal} & \text{if } (v_i, v_j) \in e_{temporal} \\ w_{positive\_semantic} & \text{if } (v_i, v_j) \in e_{positive\_semantic} \\ w_{negative\_semantic} & \text{if } (v_i, v_j) \in e_{negative\_semantic} \\ w_{self-loop} & \text{if } (v_i, v_j) \in e_{self-loop} \\ 0 & \text{if } (v_i, v_j) \notin \mathcal{E} \end{cases} \quad (12)$$

In a sense, our objective is to learn the following mapping function $\mathcal{F}$ [19, 23]:

$$\mathcal{F}: \mathcal{D}i\mathcal{G}(\mathcal{V}, \mathcal{E}) \to \mathcal{Y} = \{y_t\}_{t=1}^T \quad (13)$$

To this end, we use a Graph Neural Network (GNN) model $\mathcal{F} = f(X, \mathcal{A})$ as the backbone model to fuse the multi-modal features of nodes. Specifically, it is a two-layer graph convolutional neural network (GCN), which is a spatial-based model suitable for processing directed graphs [17, 18].

Since undirected graphs are a special case of directed graphs, the undirected graph convolution operator does not work with directed graphs. In a directed graph, the degree of a node is divided into out-degree and in-degree. Directed graphs need to perform convolution operations on them separately. Chung [24] gives the definition of Laplacian for a directed graph as follows

$$L = I_T - \frac{\Phi^{\frac{1}{2}} \mathcal{P} \Phi^{\frac{-1}{2}} + \Phi^{\frac{-1}{2}} \mathcal{P}^* \Phi^{\frac{1}{2}}}{2} \quad (14)$$

where $I_T$ is the identity matrix, $\Phi$ is a diagonal matrix, $\mathcal{P}$ is a transition probability matrix and $\mathcal{P}^* = \mathcal{P}^T$ is the conjugated transpose matrix [18, 24]. $\Phi(v_i, v_i) = \phi(v_i)$ and $\sum_{v_i} \phi(v_i) = 1$. $\phi = \frac{1}{N}$ the Perron vector of $\mathcal{P}$ [24]. $\mathcal{P}$ for a weighted directed graph is defined as [24]

$$\mathcal{P}(v_i, v_j) = \frac{w_{i,j}}{\sum_k w_{i,k}} \quad (15)$$

where $k$ is the number of neighbor nodes of $v_i$.

According to the Laplace formula mentioned above, Li et al. [18] described in detail the directed graph convolution operator considering the out-degree and in-degree, as follows

$$D_{DG}(\mathcal{A}) = \frac{\widetilde{D}_{out}^{-\frac{1}{2}} (\widetilde{\mathcal{A}} + \widetilde{\mathcal{A}}^T) \widetilde{D}_{out}^{-\frac{1}{2}}}{2} \quad (16)$$

$$D_{DG}(\mathcal{A}^T) = \frac{\widetilde{D}_{in}^{-\frac{1}{2}} (\widetilde{\mathcal{A}}^T + \widetilde{\mathcal{A}}) \widetilde{D}_{in}^{-\frac{1}{2}}}{2} \quad (17)$$

where $\widetilde{\mathcal{A}} = \mathcal{A} + I_T$ [18]. $\widetilde{D}_{out}$ and $\widetilde{D}_{in}$ are the out-degree matrix and in-degree matrix, respectively. $\widetilde{D}_{out}(i,i) = \sum_j \widetilde{A}_{ij}$ and $\widetilde{D}_{in}(j,j) = \sum_i \widetilde{A}_{ij}$.

In summary, our directed graph forward model has the following form:

$$H^{(0)} = \sigma_1(\text{norms}(\frac{D_{DG}(\mathcal{A}) X W_1^{(0)} + D_{DG}(\mathcal{A}^T) X W_2^{(0)}}{2})) \quad (18)$$

$$H^{(0)} = \text{dropout}(H^{(0)}) \quad (19)$$

$$H^{(1)} = \sigma_2(\text{MLP}(\frac{D_{DG}(\mathcal{A}) H^{(0)} W_1^{(1)} + D_{DG}(\mathcal{A}^T) H^{(0)} W_2^{(1)}}{2})) \quad (20)$$

where $X$ is the multi-modal feature matrix of node attributes. $W_1^{(0)}$, $W_2^{(0)}$, $W_1^{(1)}$ and $W_2^{(1)}$ are the weight matrices of the network layers. $\sigma_1$ is a Leaky ReLU activation function. $\sigma_2$ is a log-softmax activation function. In the first layer network, a 1D batch normalization function and dropout function are used. In the second layer network, an MLP function composed of linear units, dropout, and linear units is used.

Our model is optimized by node classification loss $L_{cls}$ and edge alignment loss $L_{edge}$. $L_{cls}$ is a cross-entropy loss [23], which computes the discrepancy between the predicted node action $\hat{y}_i$ and the ground-truth $y_i$. The formula is as

$$L_{cls} = - \sum_{i=1}^{N_{batch}} \sum_{c=1}^{C} y_i^c \log(\hat{y}_i^c) \quad (21)$$

where $N_{batch}$ is the number of nodes in each batch of graphs. $C$ is the total number of action categories, and $c$ is the $c$-th action category in $C$. $L_{edge}$ is the Kullback-Leibler (KL) [4] divergence, which calculates the divergence of the predicted adjacency matrix $\mathcal{A}'$ and the target $\mathcal{A}$. It is formulated as follows

$$L_{edge} = \frac{1}{2} \mathbb{E}_{(\mathcal{A},\mathcal{A}') \sim \mathcal{D}i\mathcal{G}} KL(\mathcal{A},\mathcal{A}') \quad (22)$$

where $(\mathcal{A}, \mathcal{A}') \sim \mathcal{D}i\mathcal{G}$ is meant to be an adjacency matrix of a graph. The total loss function is as follows

$$L_{total} = L_{cls} + \lambda L_{edge} \quad (23)$$

where $\lambda$ is the balance factor.

# 3 Experiments

## 3.1 Datasets

To evaluate our method, we perform experiments on two action datasets of food preparation. Table 1 displays the basic information for the two benchmark datasets.

*GTEA*. Georgia Tech Egocentric Activity (GTEA) datasets consist of a first-person instructional video of food preparation in a kitchen environment. It has 28 videos with an average length of 1 minute and 10 seconds [15]. Each video is split into frame sets at 15 fps [15]. Each video has an average of 20 action instances, and each frame is annotated with 11 action categories (including background) [15]. Our method is evaluated using a 4-fold cross-validation average for this dataset [15].

*50Salads*. This dataset includes instructional videos of 25 testers preparing two mixed salads in a kitchen environment [14]. It contains 50 videos with an average length of 6 minutes and 24 seconds [14]. Each video is also split into frame sets at 15 fps. Each frame is labeled with 19 action categories (including background), and each video has 20 action instances on average [14].

**Table 1** Datasets

| Dataset | Video | Frame | | | Action |
|---|---|---|---|---|---|
| | | Max | Min | Mean | |
| GTEA | 28 | 2009 | 634 | 1321 | 11 |
| 50Salads | 50 | 18143 | 7804 | 12973 | 19 |

## 3.2 Evaluation metrics

Following previous works [3, 15, 21, 33], we used the following metrics for evaluation: node-wise accuracy (Acc.), segmental edit score (Edit), and segmental overlap F1 score. Node-wise accuracy corresponds to the frame-wise accuracy of the video, which is the most used metric in action segmentation. Segmental edit score is used to compensate for the lack of node-wise accuracy for over-segmentation. Segmental overlap F1 score is used to evaluate the quality of the prediction, with the thresholds of 0.1, 0.25, and 0.5 (F1@10, F1@25, F1@50) respectively. Additionally, we also use Top-1 and Top-5 to evaluate our method for action recognition.

## 3.3 Implementation details

For all datasets, each video is divided into a frameset at 15fps. We use I3D to extract RGB features and optical-flow of each frame as visual features. Due to computer resource constraints, a frame set constructs a graph every 500 frames sequentially sampled. The remaining less than 500 frames have also constructed a graph. Node2vec [22] is used in the directed graph to encode the neighborhood information of each node as structural features. For training data, we follow ActionCLIP [4] to encode the label text as semantic features. For the unlabeled test data, we first use ActionCLIP as the backbone model to predict the action classification for each frame as initial pseudo-labels and corresponding prompt-based semantic features. Then, our model constructs the graph structure of the test data based on the initial pseudo-labels.

In the training phase, a 2-layer GCN is used as the back-bone model, where the hidden layer dimension is 512 and the optimizer is Adam. The dimension of the node attribute matrix is 2816, of which the dimension of visual features, structural features, and semantic features are 2048, 128, and 512, respectively. The batch size is set to 8, which means that each batch has 8 graphs. Furthermore, the learning rate is 0.004, the weight delay is 5e-4, and the dropout probability is 0.5. The weight $r$ of negative semantic edges can be 0 or a very small value (such as 0.1, 0.01, ...), we set it to 0 in the experiment. The balance factor $\lambda$ is 0.1. The model has trained 30 epochs. All experiments are performed on a computer with 1 NVIDIA GeForce RTX 3090 GPU, 128G memory, Ubuntu 20.04 system, and PyTorch.

# 4 Results analysis

## 4.1 Comparison with state-of-the-art methods

On the GTEA and 50Salads datasets, we evaluate the performance of our approach and other models, including state-of-the-art visual models and graph models. As seen by Table 2, compared to video-based visual models, our approach leverages graph edge design and neighborhood message aggregation to capture long-term and short-term temporal relations in videos, rather than Transformer (such as Bridge-Prompt (Br-Prompt+ASFormer), UVAST [32], ASFormer), attention mechanism, multimodal features (e.g. MCFM [31]) or other methods (e.g. DPRN, Min-Seok Kan et al. [29], ETSN, MS-TCN++ [25]) in vision models. The results show that our approach achieves the performance of SOTA, which demonstrates that it is feasible to use graph models to learn and reason about video relations. Furthermore, although both Bridge-Prompt and our approach use prompt-based methods, we also consider structural features that reflect neighborhood information. This is the reason our approach outperforms Bridge-Prompt.

**Table 2** Comparisons with other models

| Type | Model | Year | GTEA F1@{10,25,50} | | | Edit | Acc. | 50Salads F1@{10,25,50} | | | Edit | Acc. |
|---|---|---|---|---|---|---|---|---|---|---|---|---|
| visual | MS-TCN++ [25] | 2020 | 88.8 | 85.7 | 76.0 | 83.5 | 80.1 | 80.7 | 78.5 | 70.1 | 74.3 | 83.7 |
| | BCN [26] | ECCV 2020 | 88.5 | 87.1 | 77.3 | 84.4 | 79.8 | 82.3 | 81.3 | 74.0 | 74.3 | 84.4 |
| | ASRF+HASR [27] | ICCV 2021 | 90.9 | 88.6 | 76.4 | 87.5 | 78.7 | 86.6 | 85.7 | 78.5 | 83.9 | 81.0 |
| | ETSN [28] | 2021 | 91.1 | 90.0 | 77.9 | 86.2 | 78.2 | 85.2 | 83.9 | 75.4 | 78.8 | 82.0 |
| | ASFormer [21] | arXiv 2021 | 90.1 | 88.8 | 79.2 | 84.6 | 79.7 | 85.1 | 83.4 | 76.0 | 79.6 | 85.6 |
| | Min-Seok Kan et al. [29] | 2022 | 87.1 | 84.5 | 71.8 | 79.9 | 78.3 | 79.0 | 76.8 | 69.5 | 71.1 | 83.1 |
| | BCN +SCSN [30] | ICME 2022 | 85.1 | 83.4 | 77.2 | 78.4 | **85.8** | 91.9 | 90.4 | 80.5 | **89.1** | 80.2 |
| | MCFM-V+ASFormer [31] | ICIP 2022 | 91.8 | 91.2 | 80.8 | 88.0 | 80.5 | 90.6 | 89.5 | 84.2 | 84.6 | 90.3 |
| | UVAST [32] | ECCV 2022 | 92.7 | 91.3 | 81.0 | **92.1** | 80.5 | 89.1 | 87.6 | 81.7 | 83.9 | 87.4 |
| | DPRN [33] | 2022 | 92.9 | **92.0** | 82.9 | 90.9 | 82.0 | 87.8 | 86.3 | 79.4 | 82.0 | 87.2 |
| | Br-Prompt+ASFormer [15] | CVPR 2022 | **94.1** | **92.0** | 83.0 | 91.6 | 81.2 | 89.2 | 87.8 | 81.3 | 83.8 | 88.1 |
| | DiffAct [34] | ICCV 2023 | 92.5 | 91.5 | **84.7** | 89.6 | 82.2 | 90.1 | 89.2 | 83.7 | 85.0 | 88.9 |
| | Br-Prompt+ASPnet [35] | CVPR 2023 | - | - | - | - | - | **92.7** | **91.6** | **88.5** | 87.5 | **91.4** |
| graph | Bi-LSTM+GTRM [3] | CVPR 2020 | - | - | - | - | - | 70.4 | 68.9 | 62.7 | 59.4 | 81.6 |
| | MSTCN+GTRM [3] | CVPR 2020 | - | - | - | - | - | 75.4 | 72.8 | 63.9 | 67.5 | 82.6 |
| | DTGRM [6] | AAAI 2021 | 87.8 | 86.6 | 72.9 | 83.0 | 77.6 | 79.1 | 75.9 | 66.1 | 72.0 | 80.0 |
| | GCN [5] | IROS 2022 | 81.5 | 77.5 | 60.8 | 75.6 | 66.1 | 75.1 | 72.3 | 61.0 | 67.6 | 75.1 |
| | **Semantic2Graph (Ours)** | | **95.7** | **94.2** | **91.3** | **92.0** | **89.8** | 91.5 | 90.2 | 87.3 | **89.1** | 88.6 |

For graph models, although previous studies have achieved promising results, their performance is still inferior to state-of-the-art vision models. GTRM [3] represents a graph node based on the segment-level, which loses the fine-grained relations in the segment. GCN [5] combines timestamp supervision and only considers the connections between adjacent frames. DTGRM [6] outperforms the above two methods because it captures temporal relationships in videos by stacking graph convolutional layers, and it also constructs similarity graphs to model similar action relations at different moments in videos. Our approach models the temporal and semantic relations in video through different types of edges, and we also incorporate multi-modal features into node attributes, especially the prompt-based semantic features of label text. As a result, the performance of our approach is comparable to state-of-the-art vision models.

## 4.2 Effectiveness analysis

In this paper, we generate initial pseudo-labels for test data using ActionCLIP. On the test data, Semantic2Graph constructs graphs and obtains semantic features based on initial pseudo-labels. To evaluate the effectiveness of our method, we train and test SAM(SI)-HSFFM(SI) [13], ActionCLIP and Semantic2Graph on GTEA and 50Salads, respectively. As shown in Table 3, the Top-1 scores of the visual model SAM(SI)-HSFFM(SI) on GTEA and 50Salads are 65.79% and 82.06%, respectively. The results of ActionCLIP, a prompt-based visual model, are roughly comparable to those of SAM(SI)-HSFFM(SI), 69.49% and 80.82%, respectively. Compared to the visual models, the Top-1 and Top-5 scores of our method for action recognition are improved by about 10% and 3% on GTEA, and by about 6% and 2% on 50Salads. The results show that Semantic2Graph is effective in correcting the visual model results.

**Table 3** Model performance in action recognition

| Dataset | Model | Top-1 | Top-5 |
|---|---|---|---|
| GTEA | SAM(SI)-HSFFM(SI) [13] (our impl.) | 65.79 | 95.05 |
| | ActionCLIP [4] (our impl.) | 69.49 | 98.98 |
| | Ours | 89.84 | 99.92 |
| 50Salads | SAM(SI)-HSFFM(SI) [13] (our impl.) | 82.06 | 96.73 |
| | ActionCLIP [4] (our impl.) | 80.82 | 98.44 |
| | Ours | 88.61 | 99.93 |

## 4.3 Efficiency and cost analysis

In Table 4 we present the model sizes, computational complexity (FLOPs) and inference cost (running time) of representative visual models and graph models. As you can see, the graph model has fewer parameters than the visual model. Because vision models generally capture long-term dependencies in videos by stacking neural network layers and attention layers. The graph model directly constructs long-term dependencies in videos through edges. DTGRM [6] constructs multi-level dilated temporal graphs to capture the temporal dependency in video. In addition, it also stacks multiple layers of residual graph convolution layers including S-Graph (Similarity Graph) and L-Graph (Learned Graph) for temporal dependency reasoning. This results in an increase in the computation of DTGRM.

In contrast, our Semantic2Graph model has a model size of only 0.27 M due to the use of 2 layers of GCN and 1 layer of

**Table 4** Parameters, FLOPs and run time comparison

| Type | Method | #Params (M) | FLOPs (G) | Run Time (ms/frame) |
|---|---|---|---|---|
| visual | ASFormer [21] | 1.13 | 1.92 | 1.04 |
| | Bridge-Prompt [15] | 1.05 | 1.79 | 1.06 |
| graph | DTGRM [6] | 0.73 | 4.38 | 1.20 |
| | Semantic2Graph (ours) | 0.27 | 1.25 | 0.98 |

Test data comes from GTEA.

MLP. The computational complexity of Semantic2Graph is 1.25 G, which makes the model to complete a node category inference within 1 millisecond. It is a lightweight and effective model compared to others.

## 4.4 Visualization analysis

We visualize the action segmentation results of some state-of-the-art visual models (such as ASFormer and Bridge-Prompt) and graph models (such as DTGRM and our Semantic2Graph) on GTEA to quantify the impact of semantic edges on eliminating over-segmentation errors. It is observed from the color bar results in Fig. 3 that ASFormer and Bridge-Prompt suffer from over-segmentation errors for long actions (see dashed box), while DTGRM suffers from under-segmentation (see dotted circle) and boundary bias (see solid line box). The segmentation results of our Semantic2Graph on short-duration actions have very high Intersection over Union (IoU) results with ground truth, which confirms the role of semantic edges in boundary adjustment. However, like other models, Semantic2Graph also has classification errors. The possible reason is that we split a long video into multiple independent subgraphs, resulting in loss of dependencies at the split boundary. Furthermore, there is a slight action reversal error in the results of our method. We will explore reasons and solutions in future work.

## 4.5 Ablation studies

In this section, we conduct ablation studies on the GTEA dataset to determine crucial parameters and evaluate the effectiveness of components in Semantic2Graph.

### 4.5.1 The number of hop

To capture long-term relations of videos, vision models need to consider long sequences of video frames and employ LSTM, attention mechanism, or transformer [15, 21]. These methods inevitably increase the computational cost. In contrast, Semantic2Graph utilizes node2vec to encode node neighborhood information as structural features, which contains both short-term and long-term relations in the video. The lower the number of hops, the lower the cost for Semantic2Graph to capture the long-term relations of video. We conduct experiments with different hops for Node2vec.

**Table 5** Model performance with different hop number modeling video dependencies

| Method | Hop | F1@{10,25,50} | | | Edit | Acc. |
|---|---|---|---|---|---|---|
| Bridge-Prompt [15] | 16 frame | 94.10 | 92.00 | 83.00 | 91.60 | 81.20 |
| Semantic2Graph +Node2vec | 2 | 92.96 | 91.55 | 88.73 | 86.95 | 88.41 |
| Semantic2Graph +Node2vec | 3 | **95.65** | 92.75 | **91.30** | 91.88 | 88.04 |
| Semantic2Graph +Node2vec | 4 | **95.65** | 94.20 | 91.30 | 91.98 | **89.84** |
| Semantic2Graph +Node2vec | 5 | 92.86 | 90.00 | 87.14 | 87.84 | 87.15 |

From Table 5, we observe that the performance of Semantic2Graph is comparable to Bridge-Prompt [15] (which is SOTA) when node2vec selects 3 hops to capture structural features. And when selecting 4 hops, Semantic2Graph outperforms SOTA. For visual models, however, 4 hops may only capture short-term relations. TCGL [20] selects 4 frames for short-term temporal modeling. To capture long-term relations, Bridge-Prompt [15], ActionCLIP [4], and ASFormer [21] et al. utilize longer frame sequences (16, 32, or 64). Semantic2Graph has an obvious cost advantage over visual models. The primary reason is that semantic edges establish direct links between nodes with long spans. Therefore, we consider 4 hops for our subsequent experiments.

### 4.5.2 Importance of semantic in test data

To evaluate the importance of semantic edges and semantic features in test data for our method, we conduct two experiments. The semantic edges and semantic features of the test data for the first experiment are derived from the initial pseudo-labels predicted by ActionCLIP. From Table 6, the test data without semantic edges and semantic features leads to a significant drop in the performance of the model. In particular, the edit and F1 scores were reduced by around 50% on average, and the accuracy was also reduced by 9%. Furthermore, the results comparing F1@10 with accuracy (53.72% vs 80.35%) show that the test data without semantic information leads to severe under-segmentation. This just suggests the importance of semantic edges and semantic features, and that the initial pseudo-labels are reliable for our method to process test data.

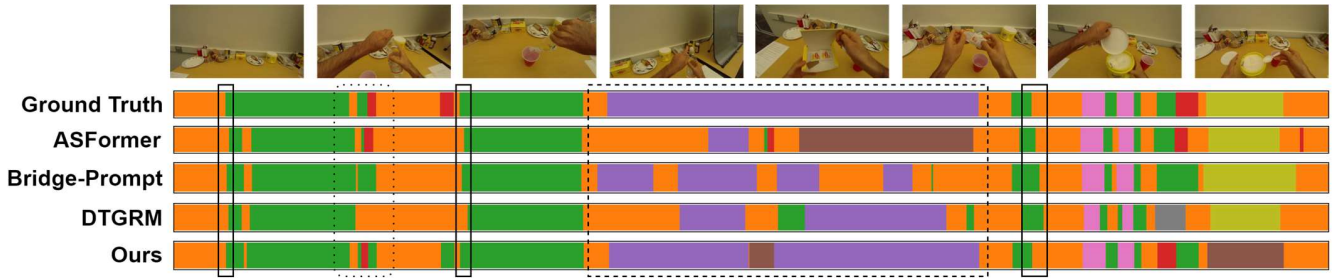

**Fig. 3** Visualization results of action segmentation on GTEA.

**Table 6** Comparing model performance on test data with and without semantic information

| Test data | F1@{10,25,50} | | | Edit | Acc. |
|---|---|---|---|---|---|
| w/ all edges and features | 95.65 | 94.20 | 91.30 | 91.98 | 89.84 |
| w/o semantic edges and features | 53.72 | 52.07 | 47.93 | 38.21 | 80.35 |

"w/" is with and "w/o" is without.

### 4.5.3 The necessity of edges

Semantic2Graph constructs the graph with three types of edges to preserve meaningful relations in video sequences. To evaluate their necessity, we performed experiments. In these experiments, the temporal edge was used as a baseline edge and was combined with the other two edges. As seen in Table 7, when self-loop edges are added to the baseline edge, the results rise slightly. Even though baseline and self-looping edges were highly accurate, other metrics, especially Edit, had lower scores. Obviously, they are under-segmentation errors [27, 28]. The reason is that the structure of the graph and the neighbors of the nodes are too monotonous, resulting in the structural features captured by node2vec being mainly short-term relations. Similarly, adding only positive or negative semantic edges also has the above results.

As shown in Table 7, the model performance improves significantly when semantic edges are added to the baseline edge. It is not difficult to find that semantic edges improve under-segmentation. Unsurprisingly, Semantic2Graph achieves the best performance over SOTA on graphs with all edges. The above results fully demonstrate the necessity of adding semantic edges and self-loop edges to capture both long-term and short-term relations cost-effectively.

To evaluate the robustness of Semantic2Graph for three edges, we also utilized a 50% random probability to add semantic edges and all edges (excluding the baseline edge). Table 7 shows that when semantic edges are added with a 50% random probability, the results drop dramatically. It is difficult for nodes lacking semantic edges to capture sufficient long-term relations when the number of hops is limited.

**Table 7** Comparing model performance for three types of edges

| Edge type | F1@{10,25,50} | | | Edit | Acc. |
|---|---|---|---|---|---|
| temporal edges (baseline) | 80.98 | 79.75 | 77.30 | 74.54 | 97.15 |
| baseline+self-loop edges | 83.54 | 82.28 | 79.75 | 77.09 | 97.19 |
| baseline+positive semantic edges | 77.46 | 76.06 | 74.65 | 70.72 | 76.26 |
| baseline+negative semantic edges | 71.26 | 66.67 | 63.22 | 59.52 | 90.45 |
| baseline+semantic edges | 89.21 | 87.77 | 80.58 | 84.13 | 73.53 |
| all edges | **95.65** | **94.20** | **91.30** | **91.98** | **89.84** |
| baseline+semantic edges (random=50%) | 81.48 | 80.25 | 77.78 | 73.78 | 88.18 |
| all edges (random=50%) | 72.53 | 72.53 | 71.43 | 69.96 | 90.15 |

### 4.5.4 The contribution of semantic features

To analyze the contribution of semantic features, we conduct some experiments in which visual and structural features serve as baseline node attributes. Table 8 illustrates the results of semantic features. For instance, using only label words and CLIP of textual prompt improves F1 scores, Edit, and Acc. by about 30% and 40%, respectively. In addition, the results indicate that CLIP is more effective than only label words. As a result of the fact that CLIP expands the label text into full sentences to acquire more robust semantic features.

**Table 8** Model performance with different semantic features

| | F1@{10,25,50} | | | Edit | Acc. |
|---|---|---|---|---|---|
| w/o semantic | 55.78 | 51.70 | 46.26 | 47.38 | 50.30 |
| only label | 88.89 | 87.50 | 81.94 | 82.59 | 78.45 |
| CLIP | 95.65 | 94.20 | 91.30 | 91.98 | 89.84 |

### 4.5.5 The efficiency of different modalities

To evaluate the effectiveness of multi-modal features, we conduct a series of experiments combining various combinations of visual, structural, and semantic modalities. The experimental results are shown in Table 9. For unimodal, semantics features with textual prompt achieve better results than visual or structural features, which indicates that high-level features are one of the keys to further improving the performance of models. Compared to unimodal, the performance of bimodal models is significantly enhanced, particularly the combination of semantic features. For multi-modal, it contains low-level, middle-level, and high-level features, achieving SOTA results [14].

**Table 9** Comparing model performance for unimodal, bimodal, and multi-modal features

| Metric | Unimodal | | | Bimodal | | | Multi-modal |
|---|---|---|---|---|---|---|---|
| | vis | str | sem | vis+str | vis+sem | str+sem | vis+str+sem |
| Top-1 | 38.6 | 13.3 | 87.3 | 50.3 | 88.8 | 85.8 | 89.8 |
| Top-5 | 90.4 | 52.3 | 89.4 | 94.8 | 99.8 | 99.7 | 99.9 |

"vis" means visual, "str" means structure, "sem" means semantic.

## 5 Related works

The significant differences between our model and previous graph-based models are as follows: First, our model incorporates additional text modalities into the node attributes to enhance semantic content, thereby improving the model's prediction accuracy. Second, positive and negative semantic edges are designed in our model to enhance the features of action segmentation boundaries.

## 5.1 Graph representation of video

Most of the prior work [21, 27] utilizes video-based visual models (such as CNN, 2D-CNNs, 3D-CNNs, VIT, etc.) to perform comprehensive video action segmentation. Vision models treat video as a sequence of RGB frames. They model complex and meaningful relations in videos through spatio-temporal feature extraction, attention mechanism, module stacking, and network depth. To obtain global or long-term relation, however, costly, and computationally intensive models are required.

Several studies suggest that transforming video into a graph-structure makes visual perception more flexible and efficient [1, 7, 19, 36]. A well-defined graph representation is critical for model performance [12]. Nodes, edges, and attributes are the fundamental elements of a graph. Common methods for obtaining nodes from a video include clip-level (snippet-level) [1, 3, 19] and frame-level [5, 6, 20] methods, among others [19, 36]. Zeng *et al.* [1] introduced a Graph Convolution Module (GCM) that constructs a graph at the snippet-level for the temporal action localization in videos. GCM extracts action units of interest from a video and represents each action unit as a node. Temporal Contrastive Graph Learning (TCGL) was proposed by Liu *et al.* [20] handle video action recognition and retrieval issues. To construct the graph at the frame-level, the video is clipped into snippets consisting of consecutive frames, and each snippet is split into frame sets of equal length. A node in the graph represents a frame in the frame set. Zhang *et al.* [19] developed a Multi-modal Interaction Graph Convolutional Network (MIGCN), which constructed a graph containing clip nodes from videos and word nodes from sentences. Regardless, clip-level methods sacrifice the video's fine-grained features in comparison to frame-level methods. In addition, other methods are only appropriate for specific task scenarios, such as MIGCN, which demands sentences to obtain word nodes. In this paper, our method adopts the frame-level to construct a graph representation of the video. Compared with clip-level, it can model more fine-grained video frame relation.

For defining edges, the temporal relation of the video is usually used as the baseline edge [1, 6, 19, 20]. In TCGL, edges are temporal prior relation (that is, the correct sequence of frames in a frame-set) [20]. Since the graph with only temporal edges loses the semantic relation implicit in videos. Therefore, some studies also introduce semantic edges [1, 19] or other edges [1, 19]. GCM added three types of edges to the graph, including contextual edges, surrounding edges, and semantic edges, to obtain contextual information, neighborhood information, and action similarity information from videos, respectively. To preserve the complex relations of video in the graph, MIGCN designs three types of edges. Specifically, the Clip-clip edge reflects the temporal adjacency relation of videos; the Word-word edge reflects the syntactic dependency between words; the Clip-word edge enhances the transmission of information between various modalities [19]. In our graph, besides temporal edges and self-loop edges, positive and negative semantic edges and weights are added to enhance the features of action segmentation boundaries. Directed edges explicitly model temporal relations, while semantic edges model action order in videos.

For attributes, the RGB feature is an essential basic attribute [1, 2, 6]. In GCM, the attributes of the nodes are the fused embedding of image features from all frames in an action unit. For TCGL, node attributes are spatial-temporal features extracted from snippets via C3D, R3D, or R(2+1)D models. Furthermore, they also obtain graph augmentation from different views by masking node features and removing edges. For MIGCN, the initialization of clip node attributes is the visual feature of the clip. Then, use BiGRU (bi-directional GRU network) to encode the semantic information of all clips in the whole video to update clip node attributes [19]. The initialization of word node attributes is the embedding of Glove, which is then encoded by BiGRU. Some studies also present multi-modal features [19, 37], which are detailed upon in Section 2.3. Our node attributes are also composed of multi-modal features, including visual, structural, and semantic features. Different from other methods, our structural features are extracted from graphs with structural prior knowledge of different video frames. Semantic features are natural language supervision signals obtained prompt-based method to enhance semantic content.

## 5.2 Multi-modal fusion

The feature representation of videos is crucial for the model to comprehensively reason the given video [19]. According to feature classification of computer vision, there are low-level, middle-level, and high-level features [14]. Low-level features are RGB features of an image [3]. It has the advantage of containing precise target locations, but the feature information is relatively discrete, such as color, edge, outline, texture, and shape. Middle-level features are spatio-temporal properties of objects in an image, such as the state or position changes of an object in time or space. High-level features are image expressed semantics that facilitates human comprehension [38], such as the meaning expressed by symbols, audio, text, object, image language descriptions, video captions, etc. Its richness in features, although it comprises coarse target locations.

In recent years, multi-modal learning has received significant attention as multi-modal features remarkable improve model performance in various tasks [2, 11, 12, 19, 37]. Audio [12], appearance [7] or depth modality information is introduced in visual tasks. Some research works [4, 12, 14, 16, 19] have shown that incorporating natural language supervision in vision tasks enhance representational power and substantially improve model performance. Similarly, multi-modal features are also introduced into graph learning tasks [11, 19, 38]. For example, Chen *et al.* [19] exploited the inter-modal interaction of objects and text to enhance model inference. Bajaj *et al.* [19] fuse phrases and visuals to augment intra-modal representations to address language grounding. In the temporal language localization problem in video, Zhang *et al.* [19] argue that in addition to sequence dependencies, the semantic similarity between video clips and syntactic dependencies between sentence words also contribute to reason video and sentence.

They used graph neural networks to explicitly model edges in the graph to learn these intra-modal relations. Lin *et al.* [11] proposed a multi-modal approach called *REMAP*, which extracts structured disease relations and text information. from multi-modal datasets to construct a complete disease knowledge graph.

Good feature fusion is very important for the performance improvement of the model. Multi-modal feature fusion methods can be classified into model-independent and model-based [12]. Model-independent methods do not directly depend on a specific machine learning, such as feature-based, decision-based and hybrid fusion methods [12]. Model-based methods deal with fusion in models, such as kernel-based models, graph models, and neural network models [12]. Among them, the advantage of the graph model is that it can utilize the structural prior knowledge of the data, which is especially suitable for modeling spatio-temporal relations. For multi-view learning, Chen *et al.* [39] propose a joint graph convolutional neural network for feature and graph fusion. Ding *et al.* [40] combined the graph neural network and CCN to form a multi-feature fusion model to solve the problem of hyperspectral image classification.

Compared to other models, our model facilitates node classification learning of video transfer graph representations to integrate the advantages of multi-modal features, especially label-text semantic features, to achieve efficient downstream video action segmentation task.

# 6 Conclusion

In this paper, we convert the challenge of video action segmentation into graph node classification utilizing a graph-based method. To preserve the fine-grained relations in the video, we construct the directed graph of the video at the frame-level and add three types of edges, including temporal, semantic, and self-loop edges. In addition to visual features, node attributes also include structural and semantic features. More critically, semantic features are the embedding of label texts based on the textual prompt. To learn multi-modal feature fusion, a GNNs model is utilized. The results of our experiments suggest that our method outperforms SOTA. Ablation experiments also confirm the effectiveness of semantic features in enhancing model performance, and semantic edges facilitate our method to capture long-term relations at a low cost.

For future work, one direction is that we continue to explore the extraction of multi-modal features and the construction of more efficient graph-structured of videos. Another direction is to apply our method to other downstream tasks, such as graph-based action verification.

# Acknowledgment

This work was supported in part by the grant from the Ministry of Science and Technology of Taiwan, under grant Nos. MOST 111-2221-E-006-115-MY2 and Nos. MOST 111-2221-E-006-160-.

# Conflict of Interests

The authors declare that they have no conflict of interest.

# Availability of data

This paper uses publicly available datasets. The download link is as follows:
GTEA: https://cbs.ic.gatech.edu/fpv/
50Salads: https://cvip.computing.dundee.ac.uk/datasets/foodpreparation/50salads/
All: https://zenodo.org/record/3625992#.Xiv9jGhKhPY

# Author contributions

Method by Junbin Zhang, Pei-Hsuan Tsai, and Meng-Hsun Tsai. Junbin Zhang wrote the model code and Pei-Hsuan Tsai contributed to the experiments. The manuscript was written by Junbin Zhang. Pei-Hsuan Tsai and Meng-Hsun Tsai reviewed the manuscript. Final manuscript read and approved by all authors.

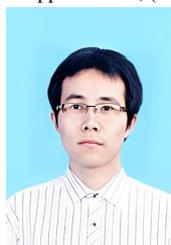
**Junbin Zhang** received the B.S. degree in optoelectronic information and the M.S. degree in optics from Fujian Normal University, China, in 2012 and 2016, respectively. From 2016 to 2019, he conducted research work at Shenzhen Graduate School, Peking University and Shenzhen Institute of Advanced Technology, Chinese Academy of Sciences. He is currently pursuing a Ph.D. degree in computer science with the Department of Computer Science and Information Engineering, National Cheng Kung University. His current research interests include graph neural networks, video understanding and computer vision.

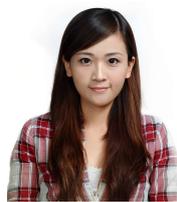
**Pei-Hsuan Tsai** received the M.Eng. degree in computer science from Cornell University, Ithaca, NY, USA, in 2004, and the Ph.D. degree in computer science from National Tsing Hua University, Hsinchu, Taiwan, in 2010. She is a Professor with the Institute of Manufacturing Information and Systems, National Chen Kung University. Her recent research interests include deep learning, computer vision, data fusion and sensor networks. She is a member of the IEEE.

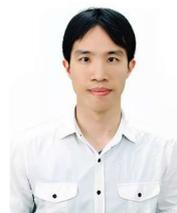
**Meng-Hsun Tsai** received the B.S., M.S., and Ph.D. degrees from National Chiao Tung University (NCTU) in 2002, 2004, and 2009, respectively. He is a Professor with the Department of Computer Science and Information Engineering, National Cheng Kung University (NCKU). He was a Visiting Scholar at University of Southern California (USC) between July and August 2012. He was a recipient of the Exploration Research Award of Pan Wen Yuan Foundation in 2012 and the Outstanding Contribution Award from IEEE Taipei Section in 2010. His current research interests include deep learning, computer vision, cybersecurity, and Internet of Things. He is a senior member of the IEEE.